\documentclass[conference]{IEEEtran}

\IEEEoverridecommandlockouts

\usepackage[T1]{fontenc}
\usepackage[utf8]{inputenc}
\usepackage{cite}
\usepackage{amsmath,amssymb,amsfonts}
\usepackage{graphicx}
\usepackage{booktabs}
\usepackage{multirow}
\usepackage{array}
\usepackage{xcolor}
\usepackage{url}
\usepackage{comment}
\usepackage{balance}
\usepackage{enumitem}
\usepackage{booktabs}
\usepackage{makecell}
\usepackage{tabularx}
\usepackage{microtype}
\usepackage{soul, xcolor}
\usepackage{xcolor}

\title{Mag-VLA: Vision-Language-Action Model for Bimanual Magnetically Actuated Microrobot Manipulation}
\author{\IEEEauthorblockN{Yongchen Wang*, Kangyi Lu*, Lan Wei, Dandan Zhang}
\thanks{*Equal contributions. All authors are with the Department of Bioengineering, Imperial-X, Imperial College London.}
}

\begin{document}
\maketitle

\begin{abstract}
Magnetically actuated microrobots have been used as wireless, non-contact manipulation tools at microscales, making them promising for minimally invasive applications. However, their control remains challenging due to indirect actuation, limited sensing, and nonlinear magnetic interactions. In this work, we propose \textbf{Mag-VLA}, a vision-language-action (VLA) model for dexterous magnetic microrobot manipulation using two robotic arms with mounted magnets for dynamic magnetic-field construction. Bimanual coordination enables capabilities such as microrobot reorientation that are difficult or infeasible with a single arm, but it also introduces coupled control challenges, as the policy must generate coordinated trajectories for both actuators within a shared workspace.
Our framework adapts a Qwen2.5-VL-7B backbone using Low-Rank Adaptation (LoRA) to process visual observations and language instructions for action prediction. To capture task progression, we introduce a motion-aware phase classifier and a phase-conditioned Action Chunking Transformer (ACT) decoder for temporally coherent multi-step control. We further construct a teleoperated magnetic microrobot manipulation dataset covering three task configurations. Ablation studies show that the ACT-based decoder substantially outperforms alternative generative action heads. In real-robot experiments, Mag-VLA achieves a 90\% approach success rate across all tasks and transport success rates of 80\%, 70\%, and 50\% as task difficulty increases. These results demonstrate that hierarchical VLA modeling provides a promising framework for magnetic microrobot manipulation.
\end{abstract}

\section{Introduction}

Magnetically actuated microrobots have attracted increasing attention in minimally invasive intervention, targeted delivery, and microscale manipulation \cite{iacovacci2024medical,zhang2023advanced, zhu2022external}. External magnetic fields enable wireless, untethered actuation without physical connections or onboard power, making these robots well suited for confined microscopic environments \cite{lin2024magnetic}. To increase control dexterity, magnetic actuation systems can further employ multiple robotic arms equipped with permanent magnets. 
However, these advantages come with substantial control challenges. Because microrobot motion is generated indirectly through magnetic interactions rather than direct contact actuation, the controller must infer task-relevant states from limited visual feedback, handle nonlinear and difficult-to-model magnetic dynamics, and maintain precise spatial reasoning in constrained workspaces. These challenges become more pronounced in the multi-actuator setting, where the policy must generate coordinated continuous trajectories for both arms in a shared workspace. Small errors in either actuator can accumulate over time, especially during transport through curved paths. Despite growing interest in autonomous control for magnetic microrobots, end-to-end frameworks that jointly integrate visual perception, language instructions, and coordinated action generation remain largely unexplored at the microscale.



\begin{figure}
\centering
\includegraphics[width=0.95\linewidth]{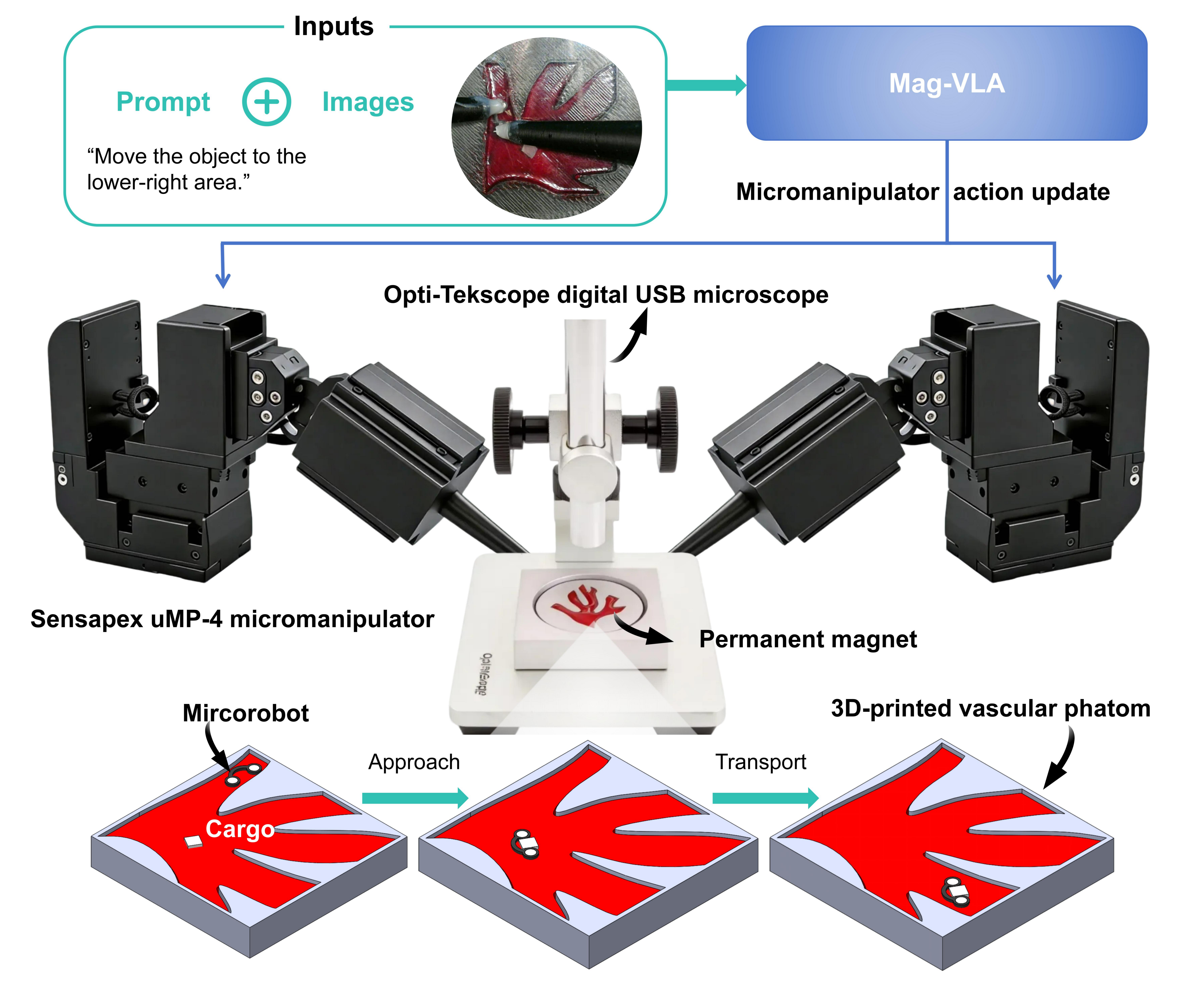}
\caption{Experimental setup and end-to-end manipulation pipeline of Mag-VLA. Two Sensapex uMp micromanipulators with permanent magnets actuate the microrobot inside a vascular phantom under top-down microscope observation. The model processes visual observations and language instructions to generate coordinated dual-arm actions across approach and transport phases.}
\label{fig:framework}
\vspace{-18pt}
\end{figure}

Recent vision-language-action (VLA) models provide a promising direction for addressing this gap. By adapting large pretrained vision-language backbones for robot control, VLA models can jointly interpret visual observations, language instructions, and action-relevant context. Representative systems such as RT-2 \cite{zitkovich2023rt}, OpenVLA \cite{kim2024openvla}, and $\pi_{0.5}$ \cite{black2025pi_} have demonstrated strong instruction following and multimodal reasoning in embodied tasks.
However, these models have primarily been developed for macro-scale settings such as tabletop manipulation, mobile interaction, and humanoid whole-body control, where action effects are visually apparent and small spatial errors are often tolerable. 

Magnetic microrobot manipulation differs fundamentally from conventional robotic manipulation. External magnetic fields actuate the microrobot, while microscope images provide only indirect estimates of its position, orientation, and interaction state. Without direct onboard measurements, the controller must infer the robot state from visual feedback alone. This is challenging because actuator motion, magnetic-field distribution, and microrobot motion are nonlinearly coupled and difficult to model accurately. In bimanual manipulation, this challenge is further amplified by the need to coordinate two magnet-mounted arms in a shared workspace. As a result, small perception or actuation errors can accumulate over time, especially during curved transport, and lead to task failure.

Recently, VLA ideas have been extended to magnetically actuated robotic systems. For example, TMR-VLA applies a VLA framework to a tri-leg silicone-based magnetic soft robot and predicts voltage commands for motion primitives such as squatting and locomotion \cite{tang2026tmr}. However, bimanual magnetic microrobot manipulation requires a different form of control. The policy must generate coordinated continuous trajectories for two magnet-mounted arms that jointly actuate the microrobot. Moreover, the manipulation process naturally consists of distinct phases, such as approach and transport, with different motion patterns and control requirements. These characteristics motivate a hierarchical VLA policy that can adapt action generation to the current manipulation phase while maintaining temporally coherent bimanual arm coordination for magnetic microrobot control.

Based on these observations, we propose \textbf{Mag-VLA}, a hierarchical VLA framework for bimanual magnetic microrobot manipulation. The overall pipeline and experimental setup are shown in Fig.~\ref{fig:framework}. Mag-VLA builds on a LoRA-adapted Qwen2.5-VL-7B backbone that jointly processes visual observations, language instructions, and robot states for action prediction. To model task progression, it uses a motion-aware phase classification head to estimate the current manipulation stage. The estimated phase then conditions an ACT decoder, which predicts temporally coherent multi-step actions for both magnetic arms \cite{zhao2023learning}. Experiments on a teleoperated dataset of 75 episodes and 20,724 RGB frames across three task configurations show that the ACT-based decoder outperforms alternative generative action heads.

The main contributions of this work are as follows:
\begin{itemize}
    \item We propose Mag-VLA, a hierarchical VLA framework that integrates visual perception, language conditioning, and coordinated dual-arm action generation for magnetic microrobot manipulation, which addresses the indirect and coupled nature of magnetic actuation without relying on explicit force or field modeling.

    \item We introduce a motion-aware phase classification head and a phase-conditioned ACT decoder, enabling phase-sensitive and temporally coherent multi-step control across approach and transport phases.

\end{itemize}



\section{Related Work}

\subsection{Autonomy in Magnetically Actuated Microrobots}
Magnetic microrobot control has been widely studied through teleoperation, where human operators guide robot motion using visual feedback and control interfaces \cite{liu2024computer}. However, bimanual magnetic manipulation increases operator workload because two actuators must be coordinated simultaneously within a shared workspace \cite{zhang2025scheduling}. This has motivated shared-control methods that combine human decision-making with robotic assistance for magnetically driven micromanipulation \cite{wang2026context}.
Toward higher autonomy, prior studies have investigated robust 3D path following, navigation in uncertain environments, and reinforcement learning for magnetic positional control \cite{tian2025automatic}. Learning-based methods have also been applied to swarm navigation and adaptive control in microrobotic systems \cite{qi2024robust, mao2025deep}. Despite these advances, most existing methods focus on tracking, navigation, or low-level magnetic control. End-to-end policies that integrate visual perception, language conditioning, and coordinated action generation remain underexplored for magnetic microrobot manipulation \cite{liu2024autonomous}.

\subsection{VLA Models for Robot Control}
A key design choice in VLA models is the action decoder. Existing approaches include direct regression and action chunking \cite{zhao2023learning}, as well as diffusion- and flow-based generative policies \cite{chi2025diffusion,lipman2022flow}. Generative decoders are effective for multimodal action distributions, but they often require iterative sampling and large-scale demonstrations. In contrast, magnetic microrobot manipulation is precision-critical, data-limited, and physically constrained. These properties motivate action-chunking formulations that favor temporally coherent and deterministic control.
Recent work has begun to apply VLA models to magnetic robotic systems. TMR-VLA predicts low-level voltage commands for a magnetically actuated tri-leg soft robot from sequential frames and language instructions \cite{tang2026tmr}. This demonstrates the potential of VLA modeling for magnetic actuation. Our work builds on this direction while focusing on bimanual microrobot manipulation, where the policy must generate coordinated continuous trajectories for two magnet-mounted arms.

\begin{figure*}[!t]
\centering
\includegraphics[width=0.85\linewidth]{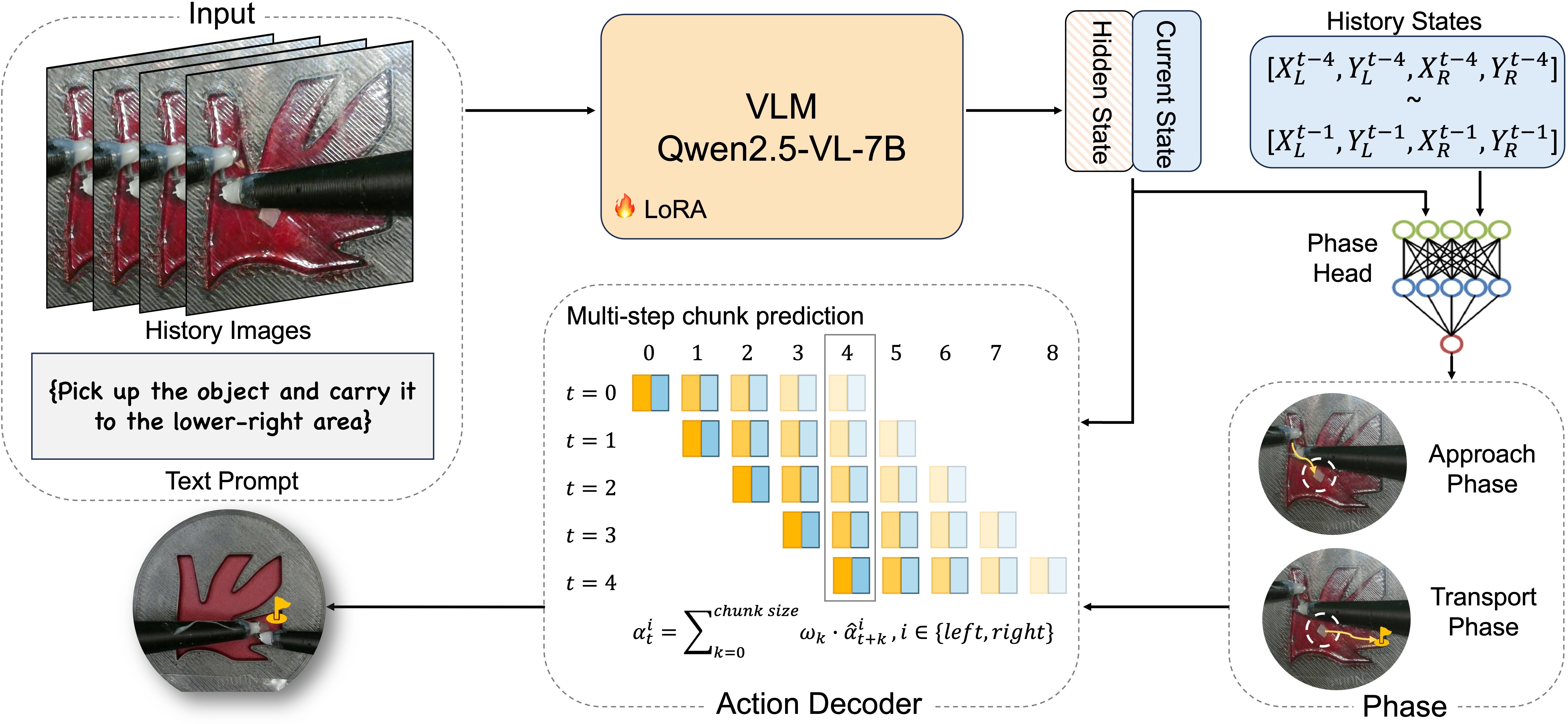}
\vspace{-5pt}
\caption{Overview of the proposed Mag-VLA framework. A history of four RGB observations and a language instruction are encoded by a LoRA-adapted Qwen2.5-VL-7B backbone. The resulting multimodal hidden states are augmented with the current robot state, while a short state history is used by a motion-aware classifier to predict the manipulation phase (approach or transport). The predicted phase token conditions an ACT-based action decoder, which generates a multi-step chunk of dual-arm control actions and applies temporal ensembling during receding-horizon execution.}
\label{fig:overview}
\vspace{-15pt}
\end{figure*}

\section{Method}


\subsection{Overview of Mag-VLA}

Mag-VLA is a hierarchical VLA policy for bimanual magnetic microrobot manipulation. As illustrated in Fig.~\ref{fig:overview}, the policy maps a history of visual observations, a language instruction, and the current dual-arm state to a future chunk of coordinated action deltas for the two magnet-mounted arms.

The model first forms a multimodal representation from visual, language, and robot-state inputs. It then estimates the current manipulation phase and uses this phase information to condition multi-step action prediction. This hierarchy allows the policy to adapt across different task stages while maintaining temporally coherent dual-arm control, without explicitly modeling magnetic forces or field distributions.

\subsection{Multimodal State Representation}

We use Qwen2.5-VL-7B as the vision-language backbone~\cite{bai2025qwen25vltechnicalreport} and fine-tune it with LoRA adapters following parameter-efficient training~\cite{hu2022lora}. At timestep $t$, the backbone takes a history of four RGB frames and a task prompt as input, and outputs multimodal hidden states
$
H_t = \{h_1,\dots,h_L\} \in \mathbb{R}^{L \times D},
$
where $L$ is the number of multimodal tokens and $D$ is the hidden-state dimension.

For coordinated bimanual control, the policy also uses the current planar positions of the two magnet-mounted arms:
$
s_t = [x_L,\, y_L,\, x_R,\, y_R] \in \mathbb{R}^{4},
$
where $(x_L,y_L)$ and $(x_R,y_R)$ denote the left and right arm positions, respectively. The state vector is projected into the backbone hidden dimension through a learned linear layer, producing a state embedding $e_t^s \in \mathbb{R}^{D}$. This embedding is appended to the hidden-state sequence:
$
\widetilde{H}_t = [H_t;\, e_t^s],
$
where $[\,;\,]$ denotes concatenation along the token dimension. The resulting conditioned memory $\widetilde{H}_t$ is used by both the phase head and the action decoder.

\subsection{Phase-Conditioned Action Prediction}
\subsubsection{Phase-Conditioned ACT Decoder}
To model task progression, Mag-VLA estimates the current manipulation phase and uses it to condition action generation. The phase head combines a masked mean-pooled summary of $\widetilde{H}_t$ with explicit motion cues, including the current normalized state and a motion feature computed from a short state-history window. A lightweight classifier then predicts phase logits,
$
\hat{p}_t \in \mathbb{R}^{2},
$
corresponding to the approach and transport phases. The predicted phase is mapped to a learned phase token $z_t^{\mathrm{phase}}$, which is prepended to the conditioned memory before action decoding.

The action decoder is a two-layer ACT module with five DETR-style learnable action queries. It performs cross-attention over the phase-conditioned memory and predicts a future chunk of five dual-arm action deltas in a single forward pass:
\begin{equation}
\hat{A}_t = f_{\mathrm{ACT}}(\widetilde{H}_t, z_t^{\mathrm{phase}}) \in \mathbb{R}^{5 \times 4},
\end{equation}
where each action step contains
$
[\Delta x_L,\, \Delta y_L,\, \Delta x_R,\, \Delta y_R].
$
Unlike diffusion- or flow-based decoders, this formulation does not require iterative sampling during inference.

\subsubsection{Inference-Time Temporal Ensembling}

During deployment, Mag-VLA follows a receding-horizon execution scheme. Each forward pass predicts an action chunk of length $K$, and consecutive replanning steps produce overlapping predictions. To obtain a smoother control signal, we apply temporal ensembling over all active chunks that cover the current timestep. If the policy predicts
$
\hat{A}^{(r)}=\{\hat{a}^{(r)}_0,\hat{a}^{(r)}_1,\dots,\hat{a}^{(r)}_{K-1}\}
$
at global step $t_r$, then for the current step $t$, the aligned index is $i=t-t_r$. The executed action is computed as
\begin{equation}
\tilde{a}_t
=
\frac{\sum\limits_{r \in \mathcal{R}_t} w_i \hat{a}^{(r)}_i}
{\sum\limits_{r \in \mathcal{R}_t} w_i},
\qquad
w_i = \exp(-\lambda i),
\end{equation}
where $\mathcal{R}_t$ is the set of active overlapping chunks and $\lambda=0.01$ is the temporal decay coefficient. The temporal buffer adds each newly predicted chunk and removes outdated ones, preserving the short-horizon structure of ACT while smoothing discontinuities between successive replans.

\subsection{Training Objective}

Mag-VLA is trained with a joint objective for action-chunk regression and phase classification. Let $K=5$ denote the action chunk length and $d_a=4$ denote the action dimension. The ground-truth action chunk at timestep $t$ is defined as
\begin{equation}
A_t = \{a_t, a_{t+1}, \ldots, a_{t+K-1}\} \in \mathbb{R}^{K \times d_a},
\end{equation}
where each action vector is
$
a_t = [\Delta x_L,\, \Delta y_L,\, \Delta x_R,\, \Delta y_R].
$
The model predicts the corresponding action chunk $\hat{A}_t \in \mathbb{R}^{K \times d_a}$.
Let $p_t \in \{0,1\}$ denote the ground-truth phase label, where the two classes correspond to the approach and transport phases. The phase head outputs logits $\hat{p}_t \in \mathbb{R}^{2}$. The overall training loss is
\begin{equation}
\mathcal{L}
=
\mathrm{SmoothL1}(\hat{A}_t, A_t; \beta=1)
+
\lambda_{\mathrm{phase}}\,\mathrm{CE}(\hat{p}_t, p_t),
\end{equation}
where $\mathrm{SmoothL1}(\cdot)$ penalizes errors in the predicted action chunk, and $\mathrm{CE}(\cdot)$ is the cross-entropy loss between the predicted phase logits and the ground-truth phase label. The parameter $\beta=1$ is the transition point of the Smooth L1 loss, and $\lambda_{\mathrm{phase}}$ balances action regression and phase classification. All action chunks are normalized during training and converted back to raw tick units during evaluation and reporting.

\section{Experiments and Results}
\subsection{Data Collection}
\begin{figure}[!t]
\centering
\includegraphics[width=1\columnwidth]{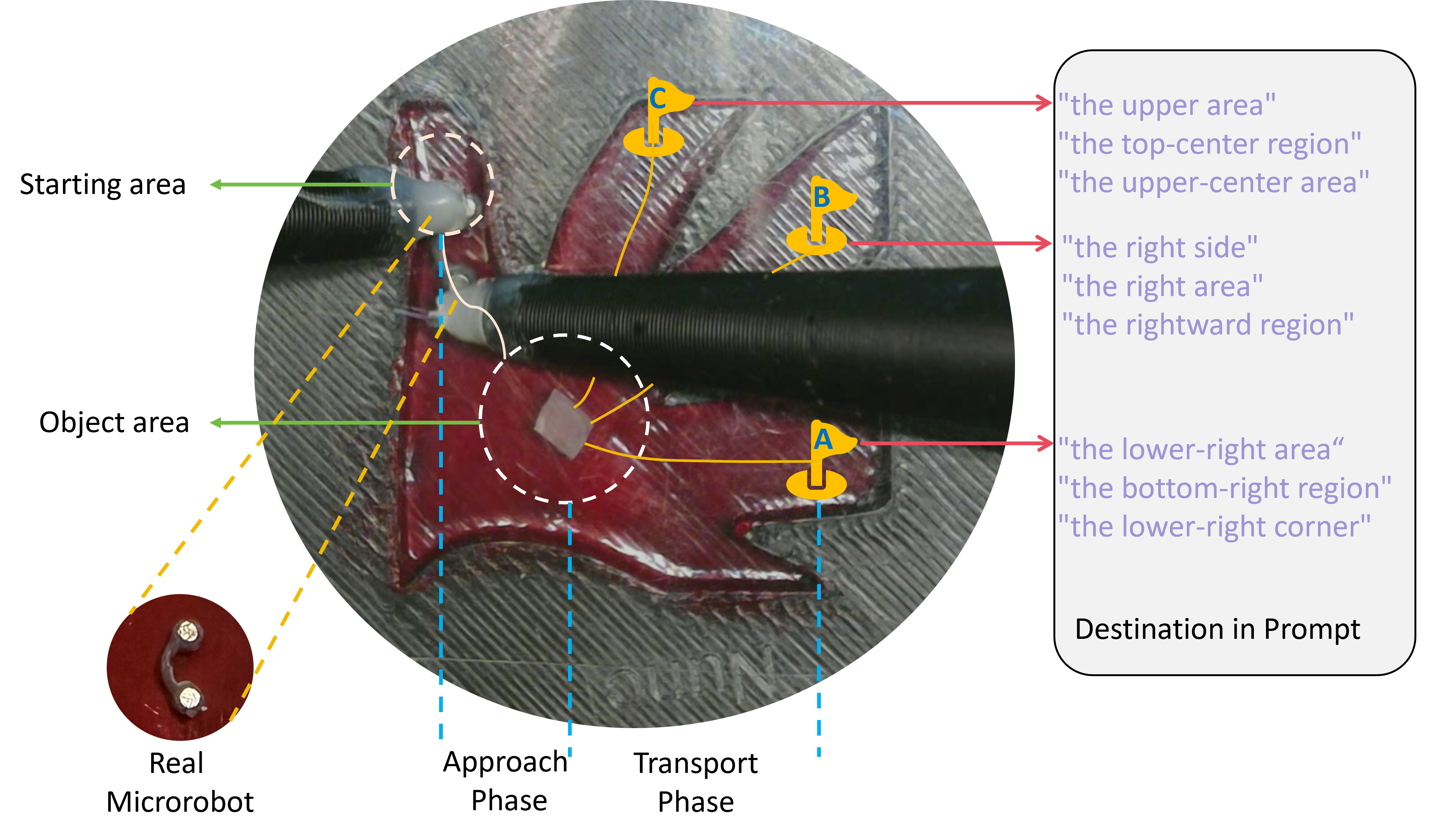}
\vspace{-25pt}
\caption{Three language-conditioned manipulation tasks (A--C) of increasing difficulty. The bimanual magnetic microrobot engages the cargo and transports it to a designated goal region. Inset: the real microrobot used in experiments.}
\label{fig:task}
\vspace{-15pt}
\end{figure}

Data are collected through teleoperation in which a human operator controls two Sensapex uMp micromanipulators by moving two Geomagic Touch haptic devices. The displacement of each haptic end effector is mapped to the translational motion of the corresponding manipulator. During teleoperation, a top-down microscope continuously captures live images of the magnetic microrobot and the surrounding workspace. These observations and synchronized robot-state data are recorded for offline training. 
The dataset contains 75 episodes captured at 1280$\times$960 resolution. The videos are downsampled from 30~Hz to 10~Hz for training, yielding a total of 20{,}724 frames. All data are organized in LeRobot format~\cite{cadene2026lerobot}. As shown in Fig.~\ref{fig:task}, we define three language-conditioned manipulation tasks, denoted Task A, Task B, and Task C, corresponding to different target regions with increasing difficulty.

In all three tasks, manipulation proceeds in two phases. During approach, the microrobot moves toward the object and establishes the interaction configuration. During transport, the object is moved to the designated target region. The dataset is split into training, validation, and test sets with a ratio of 60:9:6 episodes, yielding approximately 16.6k, 2.5k, and 1.6k samples respectively after preprocessing. To introduce a limited degree of language variation, we design a set of 70 prompts by combining different action expressions with target-region descriptions. These prompts are used to evaluate whether the policy can remain effective under modest changes in language formulation rather than a single fixed instruction template.

\subsection{Experimental Setup}
Training is performed on four NVIDIA A100 GPUs with 80 GB memory each, using an effective batch size of 64. 
All experiments use an input resolution of 448$\times$448, a 4-frame observation history, an action chunk size of 5, BFloat16 mixed-precision training, and a cosine learning-rate schedule over 5{,}000 optimization steps. For Qwen2.5-VL-7B, parameter-efficient fine-tuning is implemented with LoRA of rank 16, scale 32, and dropout 0.10. optimization uses AdamW with a weight decay of 0.01. Visual inputs are augmented at training time with random brightness scaling by a factor sampled from [0.85, 1.15] with probability 0.5 and random contrast scaling by a factor sampled from [0.90, 1.10] with probability 0.3; action labels and language prompts are kept unchanged.
Although the dataset is modest in size relative to a 7B-parameter backbone, several design choices jointly limit the effective adapted capacity on the training set. First, the Qwen2.5-VL-7B weights are kept frozen and only LoRA adapters of rank 16 with dropout 0.10 are updated, so the number of trainable parameters is several orders of magnitude smaller than the backbone; Second, AdamW weight decay of 0.01 constrains the magnitude of the adapted parameters throughout training. Third, photometric data augmentation introduces appearance-level variation that mitigates over-reliance on specific illumination cues from teleoperation sessions. Finally, the 60/9/6 episode split holds out validation and test data at the trajectory level rather than the frame level, so reported metrics reflect entirely unseen teleoperation sequences.


\subsection{Evaluation Metrics}

All quantitative results are reported on the held-out test split. We evaluate action prediction using root mean square error (RMSE), endpoint error, direction accuracy, and mean cosine similarity. RMSE is computed after de-normalizing the predicted and ground-truth action chunks, averaged over samples, prediction steps, and action dimensions. Phase-wise RMSE is computed separately on samples whose ground-truth phase labels are approach or transport. Endpoint error is the Euclidean distance between the predicted and ground-truth action vectors at the final step of the 5-step chunk. Direction accuracy is the proportion of moving samples whose predicted action has positive cosine similarity with the ground truth, while mean cosine similarity averages the same cosine values. Together, these metrics evaluate regression fidelity at the chunk level (RMSE), terminal accuracy of multi-step prediction (endpoint error), and directional consistency of executed motion (direction accuracy and cosine similarity). They jointly characterise both per-step accuracy and short-horizon trajectory quality.

\subsection{Backbone and Action-Head Comparison}

\begin{figure}[!t]
\centering
\includegraphics[width=0.95\columnwidth]{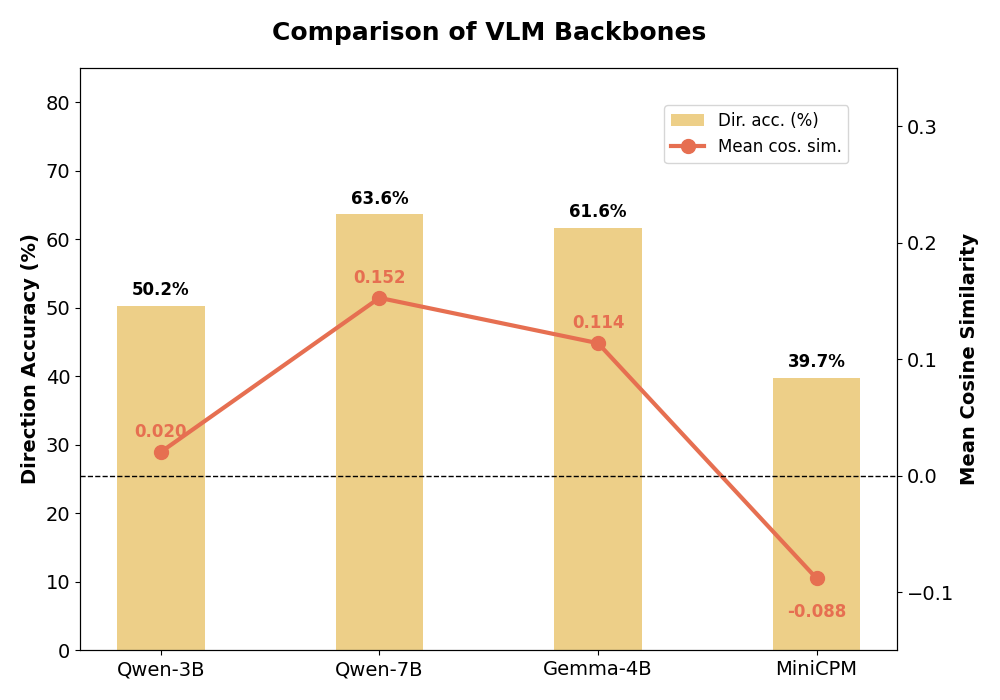}
\vspace{-15pt}
\caption{Directional comparison of VLM backbones under the shared MLP chunk head. Qwen2.5-VL-7B provides the highest direction accuracy and mean cosine similarity, indicating the better directional consistency among the tested backbones.}
\label{fig:backbone_direction}
\vspace{-14pt}
\end{figure}

We first compare four vision-language backbones under the same MLP chunk head. As shown in Table~\ref{tab:backbone_comparison}, Qwen2.5-VL-7B gives the best overall performance. It achieves the lowest overall RMSE at 107.38 ticks and the lowest approach RMSE at 106.57 ticks. Qwen2.5-VL-7B also gives the lowest endpoint mean and median errors, at 179.69 and 162.08 ticks, respectively. As shown in Fig.~\ref{fig:backbone_direction}, it further achieves the best directional performance among the tested backbones. Its direction accuracy reaches 63.59\%, and its mean cosine similarity reaches 0.1525.

\begin{table}[t]
\centering
\caption{Comparison of different VLM backbones.}
\label{tab:backbone_comparison}
\renewcommand{\arraystretch}{0.9}
\vspace{-0.15cm}
\setlength{\tabcolsep}{3pt}
{\fontsize{8}{9}\selectfont
\begin{tabular*}{\columnwidth}{@{\extracolsep{\fill}}lcccc}
\toprule
\textbf{Metric} & \textbf{Qwen-3B} & \textbf{Qwen-7B} & \textbf{Gemma-4B} & \textbf{MiniCPM} \\
\midrule
RMSE (overall) $\downarrow$        & 107.53 & \textbf{107.38} & 107.41 & 108.04 \\
RMSE (approach) $\downarrow$       & 106.72 & \textbf{106.57} & 106.78 & 107.01 \\
RMSE (transport) $\downarrow$      & 108.44 & 108.29 & \textbf{108.10} & 109.18 \\
Endpoint mean $\downarrow$         & 180.75 & \textbf{179.69} & 181.31 & 182.59 \\
Endpoint median $\downarrow$       & 162.34 & \textbf{162.08} & 164.07 & 165.35 \\
RMSE ($x_L$) $\downarrow$          & 98.79  & 97.44  & \textbf{97.40} & 99.29 \\
RMSE ($y_L$) $\downarrow$          & 122.15 & 122.13 & 122.41 & \textbf{121.82} \\
RMSE ($x_R$) $\downarrow$          & 95.46  & 95.94  & \textbf{95.06} & 97.16 \\
RMSE ($y_R$) $\downarrow$          & \textbf{111.63} & 111.84 & 112.42 & 112.03 \\
\bottomrule
\end{tabular*}
}
\vspace{1pt}
\par\footnotesize\textbf{Note:} Best results are shown in \textbf{bold}. $\downarrow$ indicates lower is better; $\uparrow$ indicates higher is better. 
\vspace{-12pt}
\end{table}

We next fix the Qwen2.5-VL-7B backbone and compare three action heads, namely ACT, Diffusion Policy, and Flow Matching, as summarized in Table~\ref{tab:chunk_comparison}. 
ACT consistently outperforms the two generative alternatives across all reported metrics. 
In particular, ACT achieves the lowest overall RMSE (79.56 ticks), compared with 153.52 for Diffusion Policy and 140.50 for Flow Matching. The same pattern holds for both the approach and transport subsets, where ACT reduces RMSE to 76.23 and 83.12 ticks, respectively.

\begin{figure}[!t]
\centering
\includegraphics[width=0.95\columnwidth]{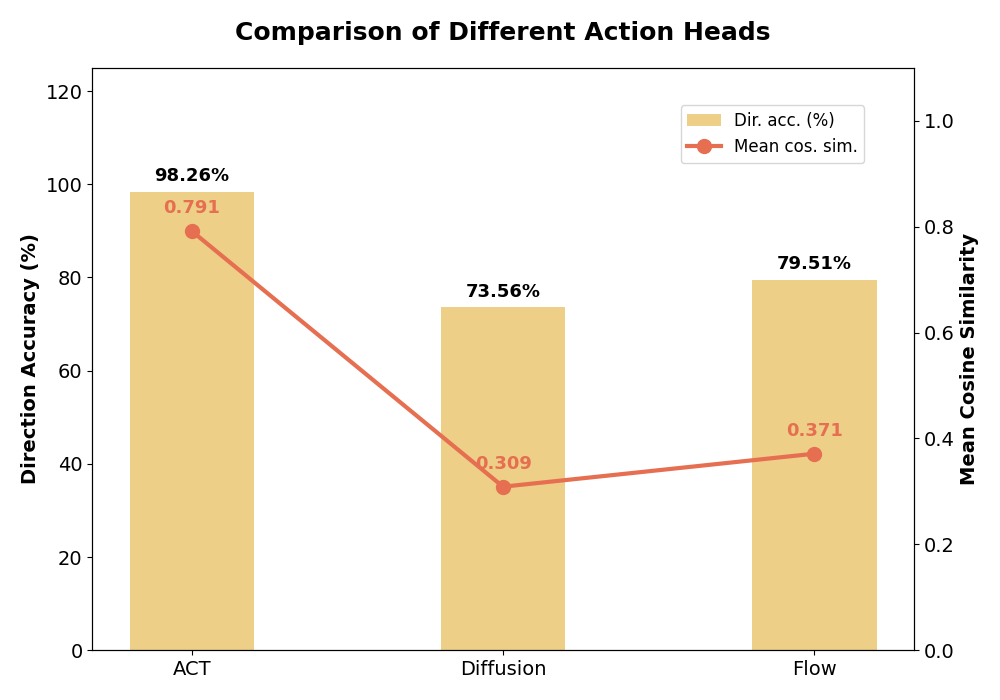}
\vspace{-15pt}
\caption{Directional comparison of different action heads with Qwen2.5-VL-7B fixed as the backbone. ACT achieves the highest direction accuracy and mean cosine similarity, indicating substantially better directional consistency than Diffusion Policy and Flow Matching.}
\vspace{-15pt}
\label{fig:chunk_direction}

\end{figure}

The advantage of ACT is also evident in endpoint and axis-wise errors. ACT attains the lowest endpoint mean and median errors, at 133.74 and 107.82 ticks, while the corresponding values roughly double to 275.96 and 260.79 for Diffusion Policy and to 265.75 and 251.25 for Flow Matching. Across all four action dimensions, ACT also gives the lowest RMSE, indicating more accurate bilateral motion prediction for both manipulators. This trend is further reflected in the directional metrics, as illustrated in Fig.~\ref{fig:chunk_direction}. ACT reaches 98.26\% direction accuracy and 0.791 mean cosine similarity, substantially outperforming Diffusion Policy and Flow Matching (Fig.~\ref{fig:chunk_direction}).
These results suggest that, for this deterministic dual-arm magnetic microrobot task, direct chunk regression is more effective than diffusion- or flow-based action generation.
We attribute this gap to a structural mismatch between generative decoders and the present task. Diffusion Policy and Flow Matching are designed to model multimodal action distributions through iterative sampling, and benefit most when observations admit multiple valid actions and large-scale demonstrations are available. Neither condition holds here: the magnetic field decouples actuation from contact, leaving an essentially unimodal action distribution per state, and dual-arm microscale demonstrations are costly to collect. Direct chunk regression instead exploits this deterministic structure and integrates naturally with the inference-time temporal ensembling used for smooth receding-horizon execution.

We further evaluate the motion-aware phase classification head on the held-out test split. As reported in the lower block of Table~\ref{tab:chunk_comparison}, the phase head achieves above 97\% overall accuracy across all three action-head variants, indicating that phase classification is a highly reliable signal in this domain and is largely decoupled from the downstream action decoder. This supports using the predicted phase token as a stable conditioning input for the action head.


\begin{table}[t]
\centering
\caption{Comparison of different action heads. Action-regression metrics (top) and phase-classification accuracy (bottom) are evaluated on the same held-out test split.}
\label{tab:chunk_comparison}
\renewcommand{\arraystretch}{0.9}
\vspace{-0.15cm}
\setlength{\tabcolsep}{3pt}
{\fontsize{8}{9}\selectfont
\begin{tabular*}{\columnwidth}{@{\extracolsep{\fill}}lccc}
\toprule
\textbf{Metric} & \textbf{ACT} & \textbf{Diffusion} & \textbf{Flow} \\
\midrule
RMSE (overall) $\downarrow$        & \textbf{79.56}  & 153.52 & 140.50 \\
RMSE (approach) $\downarrow$       & \textbf{76.23}  & 149.23 & 138.89 \\
RMSE (transport) $\downarrow$      & \textbf{83.12}  & 158.19 & 142.28 \\
Endpoint mean $\downarrow$         & \textbf{133.74} & 275.96 & 265.75 \\
Endpoint median $\downarrow$       & \textbf{107.82} & 260.79 & 251.25 \\
RMSE ($x_L$) $\downarrow$          & \textbf{76.64}  & 134.41 & 135.13 \\
RMSE ($y_L$) $\downarrow$          & \textbf{89.55}  & 182.75 & 157.55 \\
RMSE ($x_R$) $\downarrow$          & \textbf{72.72}  & 132.55 & 131.47 \\
RMSE ($y_R$) $\downarrow$          & \textbf{78.33}  & 158.88 & 136.35 \\
\midrule
Phase Acc.\ (overall) $\uparrow$   & 97.21\% & 97.15\% & 97.66\% \\
Phase Acc.\ (approach) $\uparrow$  & 97.36\% & 98.92\% & 97.60\% \\
Phase Acc.\ (transport) $\uparrow$ & 97.04\% & 95.16\% & 97.72\%\\
\bottomrule
\end{tabular*}
}
\vspace{1pt}
\par\footnotesize\textbf{Note:} Best results are shown in \textbf{bold}. $\downarrow$ indicates lower is better; $\uparrow$ indicates higher is better.
\vspace{-13pt}
\end{table}

\begin{figure}[!t]
\centering
\includegraphics[width=\columnwidth]{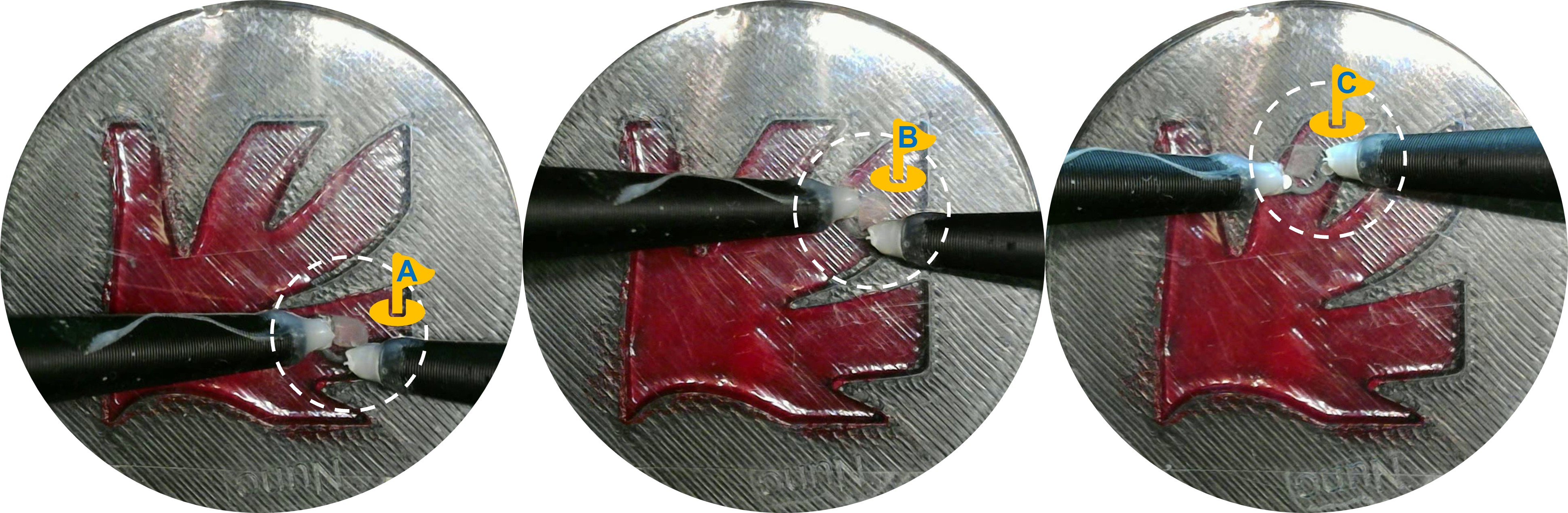}
\vspace{-15pt}
\caption{Successful real-robot demonstrations on Tasks A, B, and C. The three panels show representative final transport outcomes; the required transport path becomes progressively more curved from A to C, demanding larger reorientation of the bimanual pair.}
\label{fig:realrobot_tasks}
\vspace{-15pt}
\end{figure}

\subsection{Real-Robot Evaluation}


We finally evaluate the deployed Mag-VLA model on three real-robot tasks, each measured separately for the approach and transport stages. As shown in Fig.~\ref{fig:realrobot_tasks}, the three tasks share the same vascular phantom and bimanual setup, but they differ in the curvature of the required transport path. Task A involves a small turn with a modest cumulative turning angle. Task B requires a larger turn that introduces sustained reorientation of the bimanual pair. Task C demands the sharpest turn, with the largest cumulative turning angle along the transport path. Path curvature and the corresponding reorientation demand thus increase monotonically from A to C.

Across this difficulty progression, the model achieves 90\% approach success on all three tasks, indicating that target acquisition is largely robust to task configuration. Transport success rates decrease from 80\% on Task A to 70\% on Task B and 50\% on Task C, consistent with the increasing path curvature. Sharper turns require more aggressive mid-transport reorientation of the bimanual pair to keep the cargo aligned with the corridor, raising the precision demand on coordinated dual-arm control. Small per-step errors have more opportunities to compound during sharper turns, and the most commonly observed failure mode is cargo misalignment or slippage as the bimanual pair attempts to negotiate a large turning angle. The remaining failures therefore concentrate in object transfer rather than target acquisition, and the success-rate gradient reflects the intended difficulty progression rather than a failure on any specific task.

\section{Conclusion}
We presented Mag-VLA, a hierarchical VLA framework for bimanual magnetic microrobot manipulation, combining a LoRA-adapted Qwen2.5-VL-7B backbone with a motion-aware phase head and a phase-conditioned ACT decoder. Experiments show that ACT yields clear advantages over diffusion- and flow-based heads under identical training budgets, and real-robot evaluation confirms consistently high approach success and effective transport across tasks of increasing path curvature. These results support hierarchical VLA modeling as a practical framework for end-to-end magnetic microrobot control at microscales.

\balance
\bibliographystyle{IEEEtran}
\bibliography{references}

\end{document}